\documentclass{easychair}

\usepackage{soul}
\usepackage{url}
\usepackage[utf8]{inputenc}
\usepackage[small]{caption}
\usepackage{amsmath}
\usepackage{amssymb}
\usepackage{booktabs}
\usepackage{algorithm}
\usepackage{algorithmic}
\usepackage{stmaryrd}
\usepackage{tikz}
\usetikzlibrary{arrows.meta,positioning}
\usepackage{todonotes}
\usepackage{natbib}

\usepackage[switch]{lineno}

\def\eqref#1{equation~\ref{#1}}

\def\1{\bm{1}}

\DeclareMathAlphabet{\mathsfit}{\encodingdefault}{\sfdefault}{m}{sl}
\SetMathAlphabet{\mathsfit}{bold}{\encodingdefault}{\sfdefault}{bx}{n}

\newcommand{\R}{\mathbb{R}}

\newtheorem{theorem}{Theorem}[section]

\newtheorem{proposition}[theorem]{Proposition}

\newtheorem{definition}[theorem]{Definition}
\newtheorem{example}{Example}

\newcommand{\OMIT}[1]{}

\newif\ifdraft\drafttrue
\ifdraft
\newcommand\anthony[1]{{\color{blue}
\small [#1 - \textbf{Anthony}]}}
\newcommand\michael[1]{{\color{magenta}
\small [#1 - \textbf{Michael}]}}
\newcommand\georg[1]{{\color{brown}
\small [#1 - \textbf{Georg}]}}
\newcommand\hongjian[1]{{\color{cyan}
\small [#1 - \textbf{Hongjian}]}}

\else
\newcommand\anthony[1]{}
\newcommand\michael[1]{}
\newcommand\georg[1]{}
\newcommand\hongjian[1]{}

\newcommand\todo[1]{}
\fi

\newcommand\ints{\mathbb{Z}}
\newcommand\nats{\mathbb{N}}

\newcommand\cval{\nu}

\newcommand\ldef{:=}
\newcommand\ltbool{\mathrm{BOOL}}
\newcommand\ltint{\mathrm{INT}}
\newcommand\lpre{\mathrm{pre}}
\newcommand\lnext{\rightarrow}
\newcommand\lboolexpr{\mathit{LBExp}}
\newcommand\lintexpr{\mathit{LIExp}}
\newcommand\lcheck{\mathop{\mathrm{check}}}
\newcommand\lif{\mathop{\mathrm{if}}}
\newcommand\lthen{\mathop{\mathrm{then}}}
\newcommand\lelse{\mathop{\mathrm{else}}}
\newcommand\lmod{\mathop{\mathrm{mod}}}
\newcommand\lval{\nu}

\newcommand\leos{\#}
\newcommand\leternity{\Omega}
\newcommand\latoi{\iota}
\newcommand\tvalid{T_\forall}
\newcommand\tinclusion{T_\subseteq}
\newcommand\tequal{T_=}

\usepackage{orcidlink}

\usepackage{xcolor}

\lstdefinelanguage{Lustre}{
	keywords={node,returns,var,begin,end,if,then,else,pre},
	keywordstyle=\bfseries,
	sensitive=true,
	morecomment=[l]{--},
}

\usepackage{tcolorbox}
\tcbuselibrary{skins, breakable}
\usepackage{amsmath, amssymb}

\title{Synthesis and Verification of Transformer Programs (Technical Report)}
\titlerunning{Synthesis and Verification of Transformer Programs}

\author{
Hongjian Jiang\,\orcidlink{0009-0006-4082-2633}\inst{1}
\and
Matthew Hague\,\orcidlink{0000-0003-4913-3800}\inst{2}
\and
Philipp Rümmer\,\orcidlink{0000-0002-2733-7098}\inst{3,4}
\and
Anthony W. Lin\,\orcidlink{0000-0003-4715-5096}\inst{1,5}
}
\authorrunning{
H. Jiang
\and
M. Hague
\and
P. Rümmer
\and
A. W. Lin
}

\institute{
University of Kaiserslautern-Landau, Germany\\
\email{hongjian.jiang@rptu.de}
\and
Royal Holloway, University of London, England\\
\email{matthew.hague@rhul.ac.uk}
\and
Uppsala University, Sweden\\
\email{philipp.ruemmer@it.uu.se}
\and
University of Regensburg, Germany
\and
MPI-SWS, Germany\\
\email{awlin@mpi-sws.org}
}

\begin{document}

\maketitle

\begin{abstract}
    C-RASP is a simple programming language that was recently shown to capture 
concepts expressible by transformers.
In this paper, we develop new algorithmic techniques for automatically 
verifying C-RASPs. 
To this end, we establish a connection to the 
verification of synchronous dataflow programs in Lustre, which enables us to 
exploit state-of-the-art model checkers utilizing highly optimized 
SMT-solvers. Our second contribution addresses learning a 
C-RASP program in the first place. To this end, we provide a new algorithm 
for learning a C-RASP from examples using local search. 
We demonstrate efficacy of our implementation for benchmarks of C-RASPs in the
literature, in particular in connection to the following applications:
(1) transformer program minimization, and (2) 
constrained learning of transformer programs (based on a partial specification).

\end{abstract}

\section{Introduction}
Transformers \cite{Vaswani17} are the underlying neural network architectures behind Large
Language Models (LLMs),
using attention to aggregate information in the input sequence.
Although transformers are expressive and at the same time efficiently 
parallelizable \cite{BKLP24,len-gen-huang,YC24,YCA24,transformers-survey},
they are rather tricky to analyze. In particular, verifying transformers is 
undecidable \cite{SAL25,counting}. Moreover, equipped with Chain of Thoughts 
(i.e. scratchpads for doing intermediate
computation), they become Turing-complete \cite{turing,MS24,softmax-turing}.

To better understand the behavior of transformers, researchers developed 
declarative languages which enjoy precisely defined semantics and at the
same time faithfully capture the behavior of transformers. In 2021 Weiss et al.
\cite{weiss2021thinking} proposed the influential RASP (Restricted Access
Sequence Processing Language) paradigm.
Essentially, RASP is a programming language that operates on sequences via
operations like selection, aggregation, and simple position-wise arithmetic
operations. Over the years the RASP paradigm has been instantiated and formally
proven to model specific classes of transformers.

C-RASP (Counting RASP) \cite{YC24} is the most recent variant of RASP that has
been shown (both theoretically and experimentally) by
\cite{len-gen-huang}
to capture real-world transformers. C-RASP is essentially a refinement of
\emph{Counting Linear Temporal Logic} \cite{BKLP24}, but without features that
are not expressible by real-world transformers.
More precisely, C-RASP allows a term of the form $\#\varphi$, which counts
the number of positions $j$ to the left of the current position $i$ where the
formula $\varphi$ holds. In this way, the Dyck-1 language of balanced
parentheses over the alphabet $\{\texttt{[},\texttt{]}\}$ is expressible in
C-RASP; more on this in Section \ref{sec:pre}.
In addition, C-RASP can express languages like MAJORITY (i.e. the input string
contains more $a$'s than $b$'s), but cannot express the flip-flop language
\cite{flipflop} --- essentially, the regular language $\Sigma^*be^*$, for some
letters $b,e$ in the alphabet $\Sigma$ --- as
well as PARITY (i.e. the number of $a$'s in the input string is even), both of
which are not trainable by transformers. \cite{len-gen-huang}
showed that C-RASP is captured by length-generalizable transformers (in 
that a transformers can learn a target language, given a finite training
dataset). In particular, C-RASP is conjectured by \cite{len-gen-huang} to
capture \emph{all} length-generalizable transformer languages; this is a
formal version of the so-called \emph{RASP-L conjecture} \cite{raspl}.

Despite tight connections between C-RASP and transformers, the role of C-RASP has
mostly been to help understand the expressivity of transformers.
Here, we study the role of C-RASP as an \emph{interpretable
	surrogate model} of transformers. ``Interpretability'' is a general concept
in AI (e.g. see \cite{Krishnan2024}), but for C-RASP (which has a formally defined
semantics) it is best understood as the existence of an automatic
method to analyze C-RASP programs. In particular, this is consistent with the
term ``interpretability'' and ``explainability'' in \emph{FXAI} (Formal
eXplainable AI) \cite{MI23} and investigations in the formal aspects of AI
(e.g. \cite{OWSH20,model-interpretability,WGY24,Hong25,foundations21}).
To this end, there are two main
challenges.
The first concerns \emph{automatic verification}, since hitherto no automatic 
method to analyze C-RASP exists. In fact, checking whether 
a given C-RASP recognizes a trivial language is undecidable~\cite{yang26}.
The second concerns \emph{synthesis}, since there is no known method to 
actually learn a C-RASP program (from examples, or otherwise).

\paragraph{Contributions: }
We develop the \emph{first techniques} that address both the
verification and synthesis of C-RASP programs. We have
implemented them and shown their efficacy on
an assortment of benchmarks from the literature
\cite{len-gen-huang,yang25knee}, in relation to the following applications:
(1) transformer
program minimization, and (2) constrained learning of transformer programs.

Our first contribution is the \emph{first automatic method for verifying C-RASP
	programs}. In particular, we provide a reduction to the verification
of \emph{Lustre}, which is a synchronous datastream language with
precise syntax and semantics. Since Lustre is supported by
highly optimized model checkers (e.g. Kind2 Model Checker \cite{cmst-kind2} with
plugins to fast SMT solvers like CVC5 and Z3), we obtain an automatic method for
verifying C-RASP programs against properties like language equivalence,
inclusion, and emptiness.

Our second contribution is the \emph{first automatic method for synthesizing C-RASP
	programs} from positive and negative examples. The method exploits local search
in the form of simulated annealing. This is orthogonal to stochastic gradient
descent for training transformers, which
yields no implementable translation 
since the current translations (e.g. see \cite{YC24,yang25knee}) work only
from a specific class of idealized transformers.

We have implemented both methods and demonstrate their efficacy against two
applications: (i) C-RASP minimization (given a C-RASP, construct a smaller
C-RASP) and (ii) constrained learning of C-RASP from a partial specification.
In particular, we have tested our verifier against existing C-RASP benchmarks
in the literature (e.g. \cite{len-gen-huang,yang25knee}), and show highly
promising experimental results (see Table \ref{tab:all-results}). In summary, our tool uses for both synthesis and
verification of each benchmark a total of at most 300 seconds. In contrast,
transformer training (e.g., GPT2) using the HuggingFace \texttt{transformer} library in PyTorch for some of these benchmarks
alone could take \emph{up to several hours} to achieve the correct language!

\paragraph{Organization: } We define C-RASP and provide examples
in Section \ref{sec:pre}. In Section \ref{sec:verify}, we define Lustre and
a reduction from C-RASP to Lustre. Our synthesis algorithm is in
Section \ref{sec:sa}. Experimental results and applications are in
Section \ref{sec:eva}. We conclude with discussion and related work in Section
\ref{sec:conc}.

\section{Preliminaries}
\label{sec:pre}

Before giving the full definition of C-RASP, we begin with an example.
Let $\Sigma$ be a finite nonempty alphabet and $\Sigma^+$ denote nonempty
words\footnote{We use ``words'' and ``strings'' synonymously in the paper, as
as is standard in formal language theory.} $a_1
\ldots a_n$.
Let $\top$ and $\bot$ denote the Boolean values true and false.
C-RASP programs are given by a sequence of rules that are evaluated over each
position of an input word from $\Sigma^+$.
Rules may evaluate to Boolean or integer values, referred to as Boolean and
counting rules, respectively.
The C-RASP program below recognizes the Dyck(-1) language;
that is, the language of words over $\Sigma = \{[, ]\}$ where each $[$ is matched by an $]$ in
a well-nested fashion.
For example $[]$, $[[]]$, and $[[][]]$ are Dyck words, $[]]$ and $][$ are
not.

\begin{example} \label{ex:crasp-dyck}
	The following C-RASP program contains four rules and recognizes Dyck words.
	It is explained below.
	\[
		\begin{array}{rclcrcl}
			C_[ & \ldef & \#[ & \quad & C_] & \ldef & \#] \\
			V & \ldef & C_[ < C_] & \quad & D & \ldef & \#V = 0 \land C_[ = C_] \\
		\end{array}
	\]
	The counting rules rule $C_[$ and $C_]$ count the number of $[$ and $]$
	characters respectively.
	We use the Boolean rule $V$ to detect violations of well-nestedness.
	That is, $V$ is true at a position of the input word if the number of $[$
	characters is less than the number of $]$ characters.
	Finally, $D$ is true only at the end of a Dyck word.
	It requires that the word contains no violations of the well-nested property
	and ultimately has the same number of $[$ and $]$.
	The table below shows the full execution of the program over the word
	$[[][]]][]$ which is not a Dyck word.
	\[
		\begin{array}{cccccccccc}
				& [    & [    & ]    & [    & ]    & ]    & ]    & [    & ] \\
			C_[ & 1    & 2    & 2    & 3    & 3    & 3    & 3    & 4    & 4 \\
			C_] & 0    & 0    & 1    & 1    & 2    & 3    & 4    & 4    & 5 \\
			V   & \bot & \bot & \bot & \bot & \bot & \bot & \top & \bot & \bot \\
			D   & \bot & \bot & \bot & \bot & \bot & \top & \bot & \bot & \bot \\
		\end{array}
	\]
	A word is accepted if the value of the last rule ($D$) is $\top$ at the final
	position of the word.
\end{example}

In the sequel, we follow closely the definition of C-RASP from
\cite{len-gen-huang}, which also allows local and periodic positional encodings.
\begin{definition}[C-RASP]
	A \emph{C-RASP} program $P$ over a finite alphabet $\Sigma$ is a finite
	sequence of rules
    $P = (R_1,\dots,R_k)$
	where each $R_i$ is either a \emph{Boolean rule} or a \emph{Count rule}
	of the form $B \ldef \mathit{BExp}$ or $C \ldef \mathit{CExp}$
	respectively. $B$ and $C$ are the names of the rules.
	To recognize languages, $R_k$ is a Boolean rule that determines acceptance.
\end{definition}

\begin{definition}[Boolean Expressions]
	\label{def:bool}
  The syntax of Boolean expressions is given by the following grammar,
  in which $a \in \Sigma$ is an atomic letter predicate, $B$ is the
  name of a Boolean rule, $\diamond \in \{\land,\lor\}$ denotes
  Boolean conjunction and disjunction, $\bowtie\mbox{} \in \{=,\le,<\}$
  denotes a comparison between counting expressions, and
  $o < m \in \nats$:
	\[
	\begin{aligned}
		\mathit{BExp} ::= {} & \top \mid \bot \mid a \mid B
		\mid \neg\,\mathit{BExp}
		\mid \mathit{BExp} \;\diamond\; \mathit{BExp} \\
		& \mid \mathit{CExp} \;\bowtie\; \mathit{CExp} \mid m\%o.
	\end{aligned}
      \]
    \end{definition}
    The semantics of Boolean expressions is as expected; in
    particular, $B$ evaluates to the value of the referenced rule and
    $m\%o$ is true at positions $j$ satisfying $j \lmod m = o$.

\begin{definition}[Counting Expressions]
	\label{def:count}
  The syntax of counting expressions is defined inductively as
  follows, where $k \in \mathbb{N}$, $C$ is the name of a counting
  rule, and $r_s < r_e \in \nats$:
	\[
	\begin{aligned}
		\mathit{CExp} ::= {} &
		k \mid C
		\mid \#(\mathit{BExp})
		\mid \#_{r_s,r_e}(\mathit{BExp}) \\
		& \mid \mathit{CExp} + \mathit{CExp}
		\mid \mathit{CExp} - \mathit{CExp} \\
		& \mid \min(\mathit{CExp}, \mathit{CExp})
		\mid \max(\mathit{CExp}, \mathit{CExp}) \\
		& \mid \mathit{BExp} \;?\; \mathit{CExp} : \mathit{CExp}.
	\end{aligned}
	\]
\end{definition}
Here, at position $j$, $\#(\mathit{BExp})$ denotes the number of
positions satisfying the Boolean expression at or before position $j$.
The expression $\#_{r_s,r_e}(\mathit{BExp})$ is a variant that
only counts positions occurring between $r_s$ and $r_e$ positions
before the current position.
The conditional expression $e_b\ ?\ e_1 : e_2$ evaluates to $e_1$ if
the Boolean expression $e_b$ holds and to $e_2$ otherwise.  Constants
$k \in \mathbb{N}$ denote fixed integer values and $C$ evaluates to
the value of the referenced rule.

Counting expressions thus provide numerical summaries of Boolean rules, which
can be compared in Boolean expressions to form complex logical
conditions.
The expressions $m\%o$ and $\#_{r_s,r_e}$ are extensions of C-RASP to
handle $\mathsf{LOCAL}$ and $\mathsf{PERIODIC}$ positional 
predicates~\cite{len-gen-huang}.

C-RASP programs are evaluated over words.
Given a word $w = a_1 \ldots a_n \in \Sigma^+$ we inductively define the value
of expressions.
Let each rule be of the form $R_i = V \ldef e_V$,
that is, $e_V$ is the expression associated to the rule named $V$.
We define $\cval_w : (\mathit{BExp} \times \{1, \ldots, n\} \rightarrow
\{\top, \bot\}) \cup (\mathit{CExp} \times \{1, \ldots, n\}
\rightarrow \ints)$ to be the value of the expression at position $j$ in the
word.
The full definition is given in the Appendix \ref{app:crasp}  (e.g.\ $\cval_w(e_1 \land e_2, j) \ldef \cval_w(e_1, j) \land \cval_w(e_2, j)$).
Here we show only a selection of interesting cases.
\[
	\begin{aligned}
		\cval_w(a, j) & \ldef
			\top\ \text{if $a = a_j$ and }
			\bot\ \text{otherwise} \\
		\cval_w(B, j) & \ldef \cval_w(e_B, j) \\
		\cval_w(m\%o, j) & \ldef j \lmod m = o \\
		\cval_w(C, j) & \ldef \cval_w(e_C, j) \\
		\cval_w(\# e, j) & \ldef \sum_{i \leq j} \cval_w(e\ ?\ 1 : 0, i) \\
		\cval_w(\#_{r_s, r_e} e, j)
			& \ldef \sum_{j - r_e \leq i \leq j - r_s} \cval_w(e\ ?\ 1 : 0, i) \\
		\cval_w(e_1\ ?\ e_2 : e_3, j) & \ldef \begin{cases}
			 \cval_w(e_2, j) & \text{if $\cval_w(e_1, j) = \top$} \\
			 \cval_w(e_3, j) & \text{otherwise.}
		\end{cases}
	\end{aligned}
\]
A word $w = a_1 \ldots a_n \in \Sigma^+$ is accepted by a C-RASP program $(R_1, \ldots,
R_k)$ if $\cval_w(B, n) = \top$ where $R_k = B \ldef e$. [Remark: for
simplicity, we do not allow an empty input, but this is easy 
to handle, e.g., by allowing an end-of-string symbol (see \cite{YC24}).]

\section{Verification of C-RASP Programs}
\label{sec:verify}

We verify properties of C-RASP programs via a translation to the
synthronous dataflow language Lustre~\cite{cphp-lustre}, which is
tailored to implementing embedded systems.
Properties of Lustre programs can be checked by highly optimized
tools such as Kind2~\cite{cmst-kind2}.

\subsection{Lustre}

We do not present Lustre in full, but show a fragment
that is sufficient to show how C-RASP can be simulated.
We will begin with an example to show how Lustre programs can recognize Dyck
words.
The program operates in a similar way to the C-RASP program of
Example~\ref{ex:crasp-dyck}.

In the full definition, a Lustre program contains a set $N_1, \ldots, N_n$ of
nodes.
For simplicity of presentation we will use a single node.
Intuitively, nodes can be used to define functions that are reusable and aid
readability.
Each node has input, output, and local variables.
Our node will have one input variable (the input word) and no output variables.
We use two types: $\ltbool$ and $\ltint$.
Each variable represents an infinite stream of values of the declared type and
will broadly correspond to a single C-RASP rule.

Local and output variables are defined using standard
mathematical and Boolean combinations of variables and constants.
For example $X \ldef \neg Y$ means that at position $i$ in the stream, $X$
takes the negation of the value of $Y$ at position $i$.
The $\lpre$ operator accesses the value of a variable at the previous position.
The $\lnext$ operator can be used to replace the initial value of a stream.
A typical example of these two operators is $X \ldef 1 \lnext \lpre(X) + 1$,
which defines the sequence $1, 2, 3, \ldots$ where $1$ replaces the initial
(undefined) value of $\lpre(X) + 1$.
Kind2 supports a $\lcheck$ keyword to verify that a given Boolean expression
evaluates to true at all positions.

\begin{example} \label{ex:lustre-dyck}
	The following Lustre program recognizes Dyck words.
	It has an input variable $I$ of type $\ltint$ and local variables $N_l$,
	$C_l$, $N_r$, $C_r$, $N_V$, and $C_V$ of type $\ltint$ and $V$ and $D$ of type
	$\ltbool$.
	$D$ is designated as an output variable.
	For this example we will make the simplifying assumption that $I$ only
	takes integer values $0$ or $1$ (corresponding to $[$ and $]$).
	Our full encoding will need to enforce this assumption and use end of
	stream markers.
	The equations defining each variable are shown below and then
	explained.
\begin{lstlisting}
node DyckCounters(I: int) returns (D: bool);
var Nl, Nr, Cl, Cr, Nv, Cv : int; V : bool;
let
  Nl = if I = 0 then 1 else 0;
  Nr = if I = 1 then 1 else 0;
  Cl = (0 -> pre(Cl)) + Nl;
  Cr = (0 -> pre(Cr)) + Nr;
  V  = Cl < Cr;
  Nv = if V then 1 else 0;
  Cv = (0 -> pre(Cv)) + Nv;
  D  = (Cv = 0) and (Cl = Cr);
tel
\end{lstlisting}

	The equations $N_l$, $N_r$ and $N_V$ convert Boolean values into an integer.
	$C_l$, $C_r$ then count how many times the input was $[$ or $]$.
	$V$ is true whenever the property $C_l \geq C_r$ is violated and $C_V$
	counts how many times a violation has occurred.
	Finally $D$ is true at all positions that have not seen any violations in
	the past and are currently well-balanced.
	The table below shows an execution over the infinite stream
	$00101000\ldots$ that represents $[[][]]]]]\ldots$.
	The value of $D$ is $\top$ when the prefix of the stream is a Dyck word.
	\[
	\begin{array}{cccccccccc}
		& 0    & 0    & 1    & 0    & 1    & 1    & 1    & \ldots \\
		N_l & 1    & 1    & 0    & 1    & 0    & 0    & 0    & \ldots \\
		N_r & 0    & 0    & 1    & 0    & 1    & 1    & 1    & \ldots \\
		C_l & 1    & 2    & 2    & 3    & 3    & 3    & 3    & \ldots \\
		C_r & 0    & 0    & 1    & 1    & 2    & 3    & 4    & \ldots \\
		V   & \bot & \bot & \bot & \bot & \bot & \bot & \top & \ldots \\
		N_V & 0    & 0    & 0    & 0    & 0    & 0    & 1    & \ldots \\
		C_V & 0    & 0    & 0    & 0    & 0    & 0    & 1    & \ldots \\
		D  & \bot & \bot & \bot & \bot & \bot & \top & \bot & \ldots \\
	\end{array}
	\]
	Adding a $\lcheck$ statement
	$\lcheck \neg (C_r = 3) \lor D$
	would require the property that whenever three $]$ characters have been
	seen, then the word so far is a Dyck word.
	This property is true of the stream above, but not true of all input
	streams.
\end{example}

\begin{definition}[Lustre (fragment)]
    A Lustre program in our fragment is a node with:
    one input variable $I$ of type $\ltint$;
    local variables $B_1, \ldots, B_{m_b}$ of type $\ltbool$;
    local variables $C_1, \ldots, C_{m_c}$ of type $\ltint$;
    one equation for each local variable; and
    a $\lcheck$ statement.

    Let $\mathcal{B} = \{B_1, \ldots, B_{m_b}\}$ and $\mathcal{C}
    = \{I, C_1, \ldots, C_{m_c}\}$ denote the Boolean and integer variables.
    Boolean variables are defined using equations $B \ldef \lboolexpr$ and integer
    variables with $C \ldef \lintexpr$ where
    \[
        \begin{aligned}
            \lboolexpr ::=
                & \top \mid \bot \mid B \mid \neg \lboolexpr \\
                & \mid \lboolexpr \diamond_b \lboolexpr
                \mid \lintexpr \bowtie_b \lintexpr \\
                & \mid \lpre(\lboolexpr)
                \mid \lboolexpr \lnext \lboolexpr
        \end{aligned}
    \]
    with $B \in \mathcal{B}$, $\diamond_b \in \{\land, \lor\}$ and $\bowtie_b
    \in \{<, =, \leq\}$ and
    \[
        \begin{aligned}
            \lintexpr ::=
                 & k \mid C
                 \mid \lintexpr \diamond_i \lintexpr \\
                 & \mid \lif \lboolexpr \lthen \lintexpr \lelse \lintexpr \\
                 & \mid \lpre(\lintexpr)
                 \mid \lintexpr \lnext \lintexpr
        \end{aligned}
    \]
    with $C \in \mathcal{C}$, $\diamond_b \in \{+, -, \lmod\}$ and $k \in
    \ints$.
    The $\lcheck$ statement is of the form $\lcheck\ \lboolexpr$.
\end{definition}

Variables hold infinite streams of integers or Booleans.
Let $\lval : (\mathcal{B} \times \nats \rightarrow \{\top, \bot\}) \cup
(\mathcal{C} \times \nats \rightarrow \ints)$ be a valuation of the Boolean
and integer variables.
That is $\lval(V, i)$ is the value of the $i$th position of the stream assigned
to the variable $V$.
We generalize $\lval$ to expressions $e$.
The full generalization is given in the Appendix \ref{app:lustre} (e.g.\
$\lval(e_1 \land e_2, i) = \lval(e_1, i) \land \lval(e_2, i)$).
Interesting cases are given below.
\[
    \begin{aligned}
        \lval(\lpre(e), i) &= \begin{cases}
            \lval(e, i - 1) & \text{when $i > 0$} \\
            \text{undefined} & \text{otherwise}
        \end{cases} \\
        \lval(e_1 \lnext e_2, i) &= \begin{cases}
                \lval(e_1, i) & \text{if $i = 0$} \\
                \lval(e_2, i) & \text{otherwise}
            \end{cases} \\
        \lval(\lif e_1 \lthen e_2 \lelse e_3, i) &= \begin{cases}
            \lval(e_2, i) & \text{if $\lval(e_1, i) = \top$} \\
            \lval(e_3, i) & \text{otherwise} \\
        \end{cases} \\
    \end{aligned}
\]
Each equation $V \ldef e$ means $V$ takes a value such that $\lval(V, i) =
\lval(e, i)$ for all $i$.
An assertion $\lcheck e$ is violated if there exists some assignment $\lval$
consistent with the equations of the program such that for some $i$ we have
$\lval(e, i) = \bot$.
We say a Lustre program is \emph{valid} if no $\lcheck$ statement is violated
by any value of the input variable $I$.

While the semantics is largely straightforward, it is worth considering some
examples of $\lpre$ and $\lnext$.
An expression $1 \lnext 2$ defines a stream $1, 2, 2, 2, \ldots$.
However, notice that the expression $1 \lnext (1 \lnext 2)$ also defines $1, 2,
2, 2, \ldots$ as $\lnext$ only changes the first value.
To define a stream $1, 2, 3, 4, 4, 4, \ldots$ we use $1 \lnext \lpre(2 \lnext
\lpre(3 \lnext 4))$.

\subsection{C-RASP to Lustre}

We encode a C-RASP program $(R_1, \ldots, R_k)$ over an alphabet $\Sigma$
into Lustre as follows.
An outline of the translation can be seen by comparing
Example~\ref{ex:crasp-dyck} and Example~\ref{ex:lustre-dyck}.
Consider the C-RASP expression $\# V$.
This is simulated in Lustre by introducing new variables $N_V$ and $C_V$ with
\[
    \begin{aligned}
        N_V & \ldef \lif V \lthen 1 \lelse 0 \\
        C_V & \ldef N_V \lnext pre(C_V) + N_V.
    \end{aligned}
\]
The expression $\# V$ is then translated directly to $C_V$.
Most other C-RASP constructs allow almost direct translations, with some
additional subtlety for the counting operators.

We also need to bridge the gap between C-RASP operating over finite words and
Lustre over infinite streams of integers.
The translations are given below.

\subsubsection{Representing the Input Word}

We introduce two fresh symbols $\leos$ (end of stream) and $\leternity$
(eternity).
Let $\Sigma' = \Sigma \cup \{\leos, \leternity\}$.
Choose any injective map $\latoi : \Sigma' \rightarrow \ints$.
A finite word $a_1 \ldots a_n$ will be encoded by an infinite stream
$\latoi(a_1) \ldots \latoi(a_n) \latoi(\leos) \latoi(\leternity)
\latoi(\leternity) \latoi(\leternity) \ldots$.

For convenience, let $I \in S$ for a set $S \subseteq \Sigma'$ denote
$\bigvee_{a \in S} I = \latoi(a)$.
Similarly $I = a$ denotes $I = \latoi(a)$.

We introduce a new Boolean Lustre variable $B_I$ and assign it the following
Boolean expression that checks at each position $i$ if $I$ is a correct input word.
It asserts that each position encodes a character.
Then, it asserts that the first symbol is not $\leternity$ and $\leternity$ is the only character seen after $\leos$.
Let $R_I =$
\[
    B_I \ldef I \in \Sigma'
        \land \neg(I = \leternity)
            \lnext (
                \neg \lpre(I \in \{\leos, \leternity\})
                \lor
                I = \leternity
            ).
\]
We introduce a second variable $B_{\hat{I}}$ to track whether $B_I$ was false
at any point in the word.
Let $R_{\hat{I}} = B_{\hat{I}} \ldef B_I \lnext (B_I \land
\lpre(B_{\hat{I}}))$.
Finally, let $\mathcal{R}_I = \{R_I, R_{\hat{I}}\}$.

\subsubsection{Translating C-RASP Rules}

Let $(R_1, \ldots, R_k)$ be a C-RASP program.
For each Boolean rule $R_i = B \ldef e$, introduce a Lustre variable $V_B$ that
is of type $\ltbool$.
Similarly, introduce a variable $V_C$ of type $\ltint$ for each counting rule
$R_i = C \ldef e$.
For each C-RASP rule $R_i = X \ldef e$ we define $T(R_i)$ to be a set of
equations including the rule $V_X \ldef T(e)$ and possibly some auxiliary
equations.
We define $T(e)$ in full in the Appendix \ref{app:trans} material (e.g. $T(\neg e) \ldef
\neg T(e)$).
Interesting cases are shown here.
Recall that $I = a$ denotes $I = \latoi(a)$.
We have
\[
    \begin{aligned}
        T(a) & \ldef I = a
        & T(B) & \ldef V_B \\
        T(m\%o) & \ldef P \lmod m = o
        & T(C) & \ldef V_C \\
        T(\#(e)) & \ldef C_{\#(e)}
        & T(\#_{r_s, r_e}(e)) & \ldef C_{\#_{r_s, r_e}(e)}
    \end{aligned}
\]
where $P$ is a fresh integer variable encoding the current position.
It uses an auxiliary equation $P \ldef 0 \lnext \lpre(P) + 1$.
$C_{\#(e)}$, $C_{\#_{r_s, r_e}(e)}$, $C_{e_1}$, and $C_{e_2}$ are fresh
integer variables defined via the auxiliary equations.

The auxiliary equations of $T(R_i)$ for the examples shown are as follows.
For each instance of $C_{\#(e)}$ we create a fresh a integer variable $C_e$
that is 0 when $e$ is false and $1$ when $e$ is true.
We add the equations
\[
    \begin{aligned}
        C_e & \ldef \lif T(e) \lthen 1 \lelse 0 \\
        C_{\#(e)} & \ldef C_e \lnext \lpre(C_{\#(e)}) + C_e \\
    \end{aligned}
\]
For $C_{\#_{r_s, r_e}(e)}$ we use the shorthand $\lpre^i$ to define a ``safe''
$i$-fold application of $\lpre$ that evaluates to $0$ if there are fewer
than $i$ previous positions.
That is
$\lpre^0(e) = e$ and $\lpre^i(e) = 0 \lnext \lpre^{i-1}(e)$ when $i > 0$.
We also create a fresh integer variable $C_e$ defined as above.
We then add
\[
	C_{\#_{r_s, r_e}(e)} \ldef \sum_{r_e \leq i \leq r_s} \lpre^i(C_e).
\]

\subsubsection{Checking Properties of C-RASP}

We can verify a number of properties of C-RASP programs by translation to Lustre.
We show here how to check language inclusion.
A similar encoding can check equality

Given two C-RASP programs $P_1 = (R^1_1, \ldots, R^1_{k_1})$ and $P_2 = (R^2_1,
\ldots, R^2_{k_2})$ we can check whether $P_1$ only accepts words that are
also accepted by $P_2$.
We assume that $P_1$ and $P_2$ have disjoint rule names and translate $P_1$ and
$P_2$ to Lustre.
The only shared variable between the programs is the input variable $I$.
Suppose $V_1$ and $V_2$ are the variables corresponding to the output equations of
the two programs.
Notice that, in C-RASP, at least one character is needed to determine the value
of the output equations.
The following $\lcheck$ constraint asserts that after the first position,
either the input $I$ is incorrect, the end of the stream marker has not been
reached, or $P_2$ accepts whenever $P_1$ does.
\[
    \lcheck \top \lnext (
        \neg(B_{\hat{I}} \land I = \leos) \lor
            \neg \lpre(V_1) \lor \lpre(V_2)
    )
\]
If the property does not hold, Kind2 reports a witnessing trace of $I$.
This trace gives counter-example word after applying the inverse of $\latoi$
and removing $\leos$ and $\leternity$.

We define $\tinclusion(P_1, P_w)$ to be the Lustre program with the equations
$\mathcal{R}_I \cup \bigcup_i T(R^1_i) \cup \bigcup_i T(R^2_i)$ and the check
statement defined above.

\begin{proposition}
    Take C-RASP programs $P_1$ and $P_2$.
	$P_1$ accepts only words accepted by $P_2$ if
	$\tinclusion(P_1, P_2)$ is valid.
\end{proposition}

The translation from C-RASP to Lustre is polynomial except
for $\#_{r_s, r_e}$.
To translate $\#_{r_s, r_e}$ we use $pre^i(C_e)$ for each $r_e \leq i \leq r_s$.
If $r_s$ and $r_e$ are encoded in binary, the translation is exponential.
Hence, we assume that $r_s$ and $r_e$ are given in unary.
We do not need a similar restriction for ${m\%o}$.

\begin{proposition}
    Take C-RASP programs $P_1$ and $P_2$.
    Assume each $r_s$ and $r_e$ appearing in the $\#_{r_s, r_e}$ operators is
    encoded in unary.
    The Lustre program $\tinclusion(P_1, P_2)$ is of size polynomial in the size of $P_1$ and $P_2$.
\end{proposition}

\section{Synthesis of C-RASP Programs}
\label{sec:sa}

We synthesize C-RASP programs with a local search procedure based on simulated
annealing \cite{sl,Henderson2003}.
It starts from an initial program and repeatedly proposes a
neighboring program obtained by applying syntactic mutations.
Given a set of words labeled as positive and negative samples, we aim to synthesize a program that is consistent with the samples while remaining small and structurally simple.

\paragraph{State Space}
A C-RASP program over an alphabet $\Sigma$ is written as a finite sequence
$P = (R_1,\ldots,R_k)$
where each $R_i$ is either a Boolean rule or a Count rule, and
$R_k$ is the distinguished Boolean rule that determines language acceptance.

To bias the search toward a concise and interpretable model, we restrict the search to programs of a fixed \emph{program shape}. Formally, a shape is a tuple
\[
\sigma = (N_b,N_c,K),
\]
where $N_b$ and $N_c$ are the number of Boolean and Counting rules,
respectively, and $K$ bounds the range of numeric constants that may
appear in Counting expressions. Only programs that satisfy these
bounds are considered during search. Let $\mathcal{P}_\sigma$ denote
the set of all well-formed C-RASP programs over $\Sigma$ that contain
exactly $N_b$ Boolean rules and $N_c$ counting rules and use only
constants from $\{0,\ldots,K\}$.  The local search explores the finite
state space $\mathcal{P}_\sigma$.

\paragraph{Mutation Operator and Neighborhood}

Local search proceeds by repeatedly modifying the current program using a
stochastic mutation operator.
The mutation operator induces the neighborhood relation explored by simulated
annealing. The mutation operator defines a Markov transition
\[
\mathsf{Mutate} : \mathcal{P}_\sigma \to \mathcal{P}_\sigma \to \R
\]
mapping each program~$P$ to a distribution~$\mathsf{Mutate}(P)$ over
the successor programs; we require
$\mathsf{Mutate}(P)(P') \geq 0$ and
$\sum_{P' \in \mathcal{P}_\sigma} \mathsf{Mutate}(P)(P') = 1$ for all
$P, P' \in \mathcal{P}_\sigma$.  The
neighborhood of a program $P$ is defined as
\[
\mathcal{N}_\sigma(P)
=
\{\,P' \in \mathcal{P}_\sigma \mid \mathsf{Mutate}(P)(P')>0\,\}.
\]
Each neighbour differs from $P$ by a single conservative rewrite of the
right-hand side of one rule.
Although individual mutations are local, the neighborhood is sufficiently rich that any two programs of the same program shape can be connected by a finite sequence of mutations.

At each iteration, the mutator selects either a Boolean rule or a
Counting rule in a program~$P$, rewrites the right-hand side of the
rule, and removes rules that are unreachable from the final
Boolean expression $R_k$. All mutations preserve well-formedness and
acyclicity by construction.

\paragraph{Boolean Rule Mutation} Boolean rules are mutated by resampling or locally
modifying their right-hand sides. The mutation operator
only uses a non-nested fragment of expressions to bias the
search toward simple programs.  In particular, Boolean expressions are
limited to expressions in which logical connectives and negation may
be applied only to Boolean atoms, and nesting of Boolean operators is
disallowed.  This restriction ensures that the program remains simple,
while more complex behavior can still be expressed through
dependencies between expressions.

A Boolean atom occurring in the right-hand side of a Boolean rule $B_i$ may be one
of the following:
an alphabet test $a \in \Sigma$;
a reference to an earlier Boolean rule $B_j$ with $j<i$;
a position--period predicate $m\%o$, where $m \in \nats$ and $o<m$; or
a comparison $\mathit{CExp}_1 \;\bowtie\; \mathit{CExp}_2$, where $\bowtie \in
\{=,\le,<\}$.

In addition to full resampling, we allow \emph{micro-mutations} that preserve
expression depth while modifying only a local syntactic feature.
Examples include swapping $\land$ and $\lor$, toggling negation, flipping the
strictness of a comparison, or resampling a single leaf atom.
These micro-mutations define small, incremental moves in the neighborhood.

\paragraph{Counting Rule Mutation}
As with Boolean expressions, we work with a shallow fragment of the
syntax defined in Definition~\ref{def:count}. During mutation, the right-hand side of each counting rule is required to have
syntactic depth at most~$2$.
Counting expressions are therefore built from count atoms using at most one arithmetic operator or one conditional expression. A new right-hand side is either resampled entirely or modified using a
micro-mutation.
Resampled Counting expressions are drawn from the grammar
\[
CExp ::= x
\mid \mathsf{op}(x,y)
\mid \mathit{BExp} \;?\; x : y,
\]
where $x,y$ are count atoms and
$\mathsf{op} \in \{+,-,\min,\max\}$.

Count atoms are drawn from a finite, type-directed pool:
integer constants $k \in \nats$;
references to earlier counting rules $C_j$ with $j<i$; or
count of $\mathsf{PERIODIC}$ predicate $\#(\phi)$ or $\mathsf{LOCAL}$ predicate $\#_{r_s,r_e}(\phi)$.
The predicate $\phi$ is a Boolean atom that does not contain comparisons or
equalities.
For $\mathsf{LOCAL}$ predicates, bounds are normalized so that
$r_s \le r_e$.

\paragraph{Objective Function}

Each candidate program $P$ is evaluated on a fixed set of labeled examples
$D = D^+ \cup D^-$.
The primary objective is to minimize the number of misclassified samples:
\[
\mathrm{mis}(P)
=
\sum_{x\in D^+}\mathbb{I}[P(x)=0]
+
\sum_{x\in D^-}\mathbb{I}[P(x)=1].
\]
Here, $\mathbb{I}[\cdot]$ is the indicator function, equal to 1 if $P$ misclassifies $x$ and 0 otherwise. Among programs with equal classification behavior, we prefer simpler programs.
We penalize on 
expressions unreachable from $R_k$ in the dependency graph, and the total
size of the abstract syntax tree.
The resulting scoring function is
\[
E(P)
=
\lambda_{\mathrm{mis}}\cdot \mathrm{mis}(P)
+
\lambda_U \cdot \mathrm{unreach}(P)
+
\lambda_S \cdot \mathrm{size}(P),
\]
where $\lambda_{\mathrm{mis}} \gg 1$.
This demands that the search first minimizes
$\mathrm{mis}(P)$ and, among programs with equal misclassified count,
prefers structurally simpler programs.

\paragraph{Annealing Dynamics}

At each iteration, the algorithm maintains a current program $P$ and proposes a
neighbouring program $P'$.
Let $\Delta = E(P') - E(P)$.
If $\Delta \le 0$, the new program~$P'$ is always accepted.
Otherwise, it is accepted with probability
$\exp\!\left(-\frac{\Delta}{T}\right)$,
where $T>0$ is the current temperature. The temperature is initialized to $T_0$ and decreased using geometric cooling.
To mitigate premature convergence, we optionally apply periodic reheating.

The complete simulated-annealing synthesis procedure is presented in
Algorithm~\ref{alg:sa}, which maintains the current program and the
best program encountered so far, applies mutation-based proposals, and
accepts or rejects them according to the probability above.
Note that if $\Delta \le 0$, then $\exp(-\Delta/T) \geq 1$ and the
new program~$P'$ is accepted with probability~$1$ in line~7.
Termination occurs when the iteration budget is exhausted or when no
further improvement is observed under a sufficiently low
temperature.

\begin{algorithm}[htbp]
	\caption{Simulated-Annealing Synthesis for C-RASP Programs}
	\label{alg:sa}
	\begin{algorithmic}[1]
		\STATE \textbf{Input:} initial program $P_0$; sample set $D$;
		temperature parameters $(T_0,\alpha,K,\rho)$;
		iteration budget $N$
		\STATE \textbf{Output:} best program $P^\star$
		
		\STATE $P \leftarrow P_0$, $P^\star \leftarrow P$,  $T
		\leftarrow T_0$
		
		\FOR{$i=0$ to $N-1$}
		\STATE Pick $P'$ according to distribution
		$\mathsf{Mutate}(P)$
		\STATE Pick $r \in [0, 1]$ uniformly at random
		
		\IF{$r \le \exp(-(E(P')-E(P))/T)$}
		\STATE $P \leftarrow P'$
		\ENDIF
		
		\IF{$E(P) < E(P^\star)$}
		\STATE $P^\star \leftarrow P$
		\ENDIF
		
		\STATE $T \leftarrow \alpha \cdot T$
		\IF{$K>0$ and $(i+1)\bmod K = 0$}
		\STATE $T \leftarrow \rho \cdot T$
		\ENDIF
		\ENDFOR
		
		\STATE \textbf{return} $P^\star$
	\end{algorithmic}
\end{algorithm}

As a refinement (not shown in Algorithm~\ref{alg:sa}),
once a program with $\mathrm{mis}(P)=0$ is discovered, the search may continue
under a hard constraint that forbids reintroducing misclassification, while
increasing structural penalties to accelerate simplification.

\section{Experiments}
\label{sec:eva}

We implement a unified Scala framework integrating \emph{C-RASP synthesis and verification} within a single modular toolchain.
The framework --- combining example-driven local search with 
verification --- consists of three core components:
\begin{enumerate}
	\item C-RASP synthesis based on simulated annealing,
	\item a lightweight C-RASP evaluator for fast scoring on finite datasets, and
	\item a verification back-end that translates C-RASP programs to Lustre and checks properties using Kind2.
\end{enumerate}

Building on this integration, our framework supports two verification-guided synthesis applications: 1) Transformer Program Minimization and 2) Constraint Learning.
\begin{enumerate}
	\item  Given a specification program $P_{\mathit{spec}}$, the goal is to synthesize a program $P$ such that $L(P)=L(P_{\mathit{spec}})$ while minimizing structural complexity. The procedure alternates between local search and equivalence checking; when equivalence fails, Kind2 provides a counterexample trace that is added to the trace set to guide further search. The process terminates once equivalence is established or a predefined limit is reached.
	\item  Given labeled examples $D = D^+ \cup D^-$ and a partial specification program $P_{\mathit{spec}}$, constraint learning aims to synthesize a program $P$ that fits all examples and satisfies $L(P) \subseteq L(P_{\mathit{spec}})$. Local search is guided by $D$ and each candidate is checked for inclusion; if the check fails, the resulting counterexample is added to $D$ as a negative example and synthesis continues.
\end{enumerate}

We evaluate on a benchmark suite of regular, counting-based, and
context-free–style languages.
For each task, synthesis is trained on $1000$ balanced examples with string
lengths in $[\ell_{\min},100]$, where $\ell_{\min}$ is the minimum valid length
for the target language, using an 80\%--20\% train/test split to report
both train accuracy~(TA) and evaluation accuracy~(EA).
The objective function is parameterized with $\lambda_{\mathit{mis}} = 1000$,
$\lambda_U = 200$, and $\lambda_S = 100$.
Simulated annealing runs a budget of $10^5$ iterations.
The temperature (initially $T_0=1.0$) is cooled exponentially with rate
$\alpha=0.9995$, and reheated every $4000$ iterations by a factor of $1.2$. 

\begin{table*}[htbp]
	\centering
	\caption{C-RASP synthesis results across benchmark languages (timeout: 300s).
		SL synthesis reports train accuracy (TA), evaluation accuracy (EA), and runtime;
		program minimization and constraint learning report refinement rounds (R),
		synthesis time (Synth), and verification time (Verif).
		``--'' indicates that no 100\%-accurate program was found within the timeout.}
	
	\label{tab:all-results}
	{
		\renewcommand{\arraystretch}{0.75}
		\setlength{\tabcolsep}{3pt}
		\resizebox{\linewidth}{!}{%
			\begin{tabular}{lcccccc ccc}
				\toprule
				& \multicolumn{3}{c}{\textbf{SL Synthesis}}
				& \multicolumn{3}{c}{\textbf{Program Minimization}}
				& \multicolumn{3}{c}{\textbf{Constraint Learning}} \\
				\cmidrule(lr){2-4} \cmidrule(lr){5-7} \cmidrule(lr){7-10}
				\textbf{Language}
				& \textbf{TA (\%)}			& \textbf{EA (\%)} & \textbf{Time (s)}
				& \textbf{R} & \textbf{Synth (s)} & \textbf{Verif (s)}
				& \textbf{R} & \textbf{Synth (s)} & \textbf{Verif (s)} \\
				\midrule
				Dyck-1               & 99.75  & 99.5 & 3.92  & 21 & 27.7  & 41.7 & 12 & --    & --   \\[0.6ex]
				AStarBStar            & 100 & 100 & 24.62  & 11 & 7.7   & 46.8 & 1  & 10.2  & 8.9  \\[0.6ex]
				AnBnCn        & 100 & 100 & 80.05  & 22 & 88.7  & 42.4 & 2  & --    & --   \\[0.6ex]
				AAStar             & 100 & 100 &  1.57   & 4  & 1.0   & 10.0 & 1  & 15.5  & 8.3  \\[0.6ex]
				ContainsAB & 100 & 100 & 17.35 & 6  & 2.2   & 12.5 & 1  & 145.0 & 10.42 \\[0.6ex]
				Majority             & 100  & 100 & 6.02   & 9  & 4.8   & 19.9 & 1  & 61.4  & 8.9  \\[0.6ex]
				Existential          & 93.12  & 92.5  & 4.62  & 6  & 2.2   & 12.5 & 4  & 48.0  & 13.6 \\[0.6ex]
				PT-2                 & 100 & 100 & 13.30  & 12 & 9.2   & 29.7 & 1  & 133.7 & 12.0 \\[0.6ex]
				PT-3                 & 100 & 100 &  59.06 & 21 & 91.3  & 46.4 & 3  & --    & --   \\[0.6ex]
				PT-12                & 100 & 100 &  90.425 & 20 & --  &  -- & 10  & --    & --   \\[0.6ex]
				$D_2$                & 99.75 & 100 &  2.26  & 27 & 63.9  & 90.8 & 11 & 68.5  & 48.2 \\[0.6ex]
				$D_3$                & 99.00 & 98.01  & 1.54   & 12 & --    & --   & 15 & --    & --   \\[0.6ex]
				$D_{12}$                & 99.75  & 98.01  & 1.32  & 22 & --    & --   & 17 & --    & --   \\[0.6ex]
				Tomita~1             & 100 & 100 &  3.45  & 5  & 1.4   & 11.2 & 1  & 28.9  & 8.3  \\[0.6ex]
				Tomita~2             & 100 & 100 & 3.03  & 21 & 33.6  & 58.3 & 9  & 78.4  & 38.3 \\[0.6ex]
				Tomita~4             & 100 & 100 & 21.32  & 21 & 50.2  & 52.0 & 2  & --    & --   \\[0.6ex]
				Tomita~7             & 100 & 100 & 9.32   & 22 & 79.1  & 59.8 & 1  & 279.4 & 18.3 \\[0.6ex]
				Next(Argmax)         & 100 & 100&  7.20   & 27 & 153.5 & 46.9 & 2  & 185.7 & 6.3  \\[0.6ex]
				\bottomrule
			\end{tabular}%
		}
	}
\end{table*}

$\mathsf{SL\ Synthesis}$ is the backbone of both applications and achieves high
accuracy across benchmarks, including those requiring non-trivial counting
(Table~\ref{tab:all-results}). Consistent with the known expressivity theory~\cite{len-gen-huang,yang25knee}, languages beyond C-RASP’s expressive power 
(Parity, Tomita~3, Tomita~5, etc.) could not be learned with a sufficiently high
accuracy; within the alloted time, the best training accuracy is below 80\%. 
We show these results in Table \ref{tab:fail-results} and the corresponding transformer training comparisons can be
found in Table \ref{tab:comparison}.

$\mathsf{Program\ Minimization}$ and $\mathsf{Constraint\ Learning}$ are
\emph{correct by construction}: a candidate is accepted only after Kind2
discharges the equivalence (resp.\ inclusion) checked against $P_{\mathit{spec}}$, and every failed check is recycled as a counterexample. Accuracy is therefore $100\%$ on all inputs whenever the loop succeeds, so the table reports only refinement rounds and synthesis/verification time. In $\mathsf{Program\ Minimization}$, local search quickly finds correct but non-minimal programs, while verification dominates the runtime by driving structural simplification, especially for richer counting structures. In $\mathsf{Constraint\ Learning}$, refinement rounds are typically few since the labeled examples restrict the hypothesis space; verification cost stays low, while synthesis grows under tighter semantic constraints.

\section{Discussion and Related Work}
\label{sec:conc}

\paragraph{Formal models of transformers:}
The intricacy of formally modeling transformers arises 
from the following features (among
others): precision (polynomial, logarithmic, and fixed), attention 
mechanisms (hardmax/softmax attention), positional encodings
(periodic, local, relative, etc.), and uniformity (a uniform description 
independent of the input length). These gave 
rise to a number of different formal models (e.g.  see 
\cite{transformers-survey}). Over the years, experimental results 
concerning learnability revealed that several variants of softmax transformers 
\cite{len-gen-huang,yang25knee} 
that are tightly related to C-RASP capture concepts that are learnable by 
transformers. \cite{len-gen-huang} proposed a formal version of the RASP-L
conjecture \cite{raspl} that C-RASP captures \emph{precisely} languages that
are expressible by length-generalizable transformers.

\paragraph{RASPs:} \cite{FWC23} proposed learning a RASP by pre-restricting the
architecture and parameters for transformer training. The difficulty with this
approach (especially for C-RASP) is to find the right ``restriction'' so that a
C-RASP can be extracted; we are not aware of such work. Our synthesis algorithm
directly finds a C-RASP without going through a transformer, which we showed to
be competitive with transformer training on our benchmarks.

B-RASP \cite{YCA24} is a version of RASP, which supports manipulations of 
bit sequences. 
B-RASP can express languages that are not efficiently learnable 
(see \cite{len-gen-huang}), including the flip-flop language \cite{flipflop}, 
which makes it not a suitable model for real-world transformers. 
The model overapproximates concepts expressible by
fixed-precision transformers \cite{characterizing}.

\paragraph{Verification of neural networks:} Such works (e.g.
\cite{reluplex,survey-NN-verification,survey-NN-verification-liu,HKWW17})
focus mostly on
verifying neural networks with a fixed number of input nodes.
Yearly competition (e.g. see \cite{VNN24}) is now held on such verification 
tasks. Here, the verification problem is decidable (in fact, NP-complete), which
is not the case for transformers \cite{SAL25}.

\paragraph{Future Work:} 
Numerous open problems lie ahead 
including the extension of our methods to other properties, e.g., 
``robustness''. This is a common property in 
neural networks verification (e.g. \cite{VNN24}), but the right formulation 
(that is motivated by practical applications) for
sequential models is yet to be explored,
e.g., perhaps via metrics on strings like
hamming/edit distances. Since practical transformers use ``Chain of
Thoughts (CoTs)'' \cite{jiang2026softmax}, another direction is to
incorporate CoTs in our framework.

\section*{Acknowledgments}
This material is based in part upon work supported by
the European Union\footnote{
	Views and opinions expressed are however those
	of the author(s) only and do not necessarily reflect those of the European
	Union or the European Research Council Executive Agency. Neither the
	European Union nor the granting authority can be held responsible for them.}
\includegraphics[width=0.75cm]{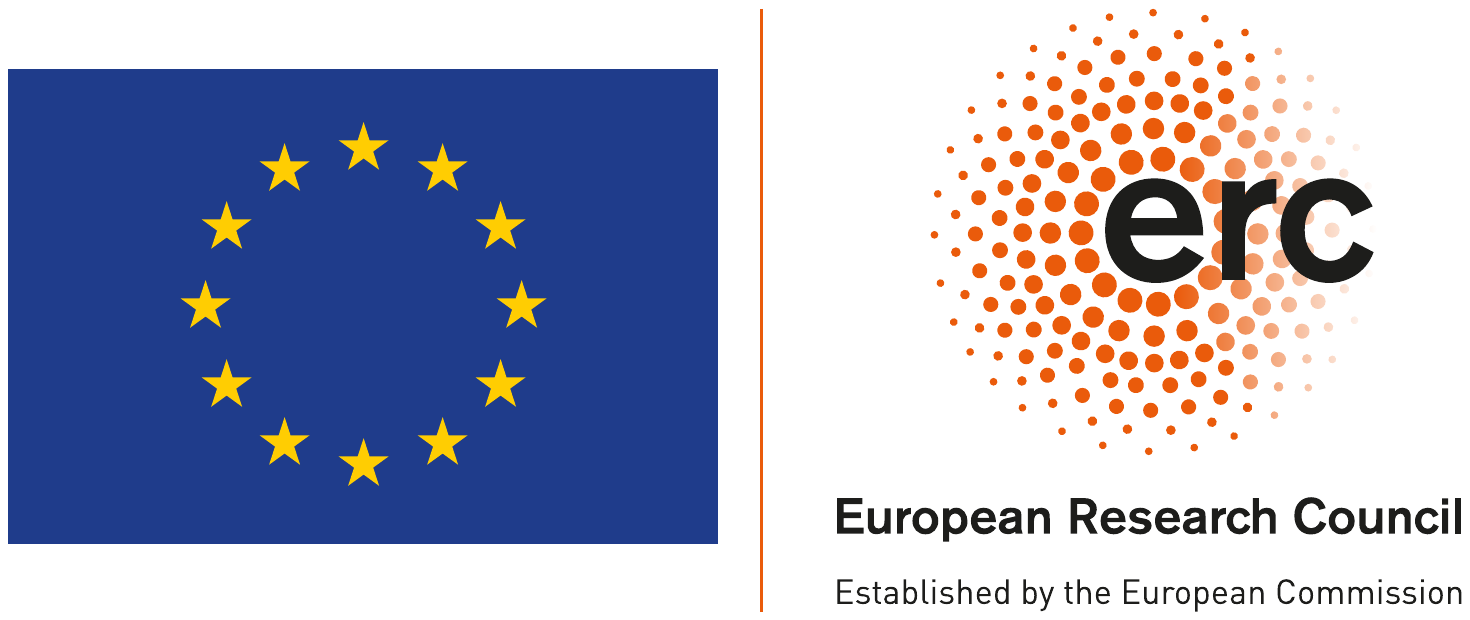}
(ERC, LASD, 101089343, \url{https://doi.org/10.3030/101089343})
and by the Swedish Research Council through grants 2021-06327 and 2025-06185.

\bibliographystyle{plain}
\bibliography{ijcai26}
\newpage
 \appendix
\section{C-RASP Semantics}
\label{app:crasp}
We give the full definition of $\cval_w(e, j)$ over a word $w = a_1 \ldots a_n$.
For Boolean expressions we have
\[
	\begin{aligned}
		\cval_w(\top, j) & \ldef \top \\
		\cval_w(\bot, j) & \ldef \bot \\
		\cval_w(a, j) & \ldef \begin{cases}
			\top & \text{if $a = a_j$} \\
			\bot & \text{otherwise}
		\end{cases} \\
		\cval_w(B, j) & \ldef \cval_w(e_B, j) \\
		\cval_w(\neg e, j) & \ldef \neg \cval_w(e, j) \\
		\cval_w(e_1 \diamond e_2, j)
			 & \ldef \cval_w(e_1, j) \diamond \cval_w(e_2, j) \\
		\cval_w(e_1 \bowtie e_2, j)
			 & \ldef \cval_w(e_1, j) \bowtie \cval_w(e_2, j) \\
		\cval_w(m\%o, j) & \ldef j \lmod m = o \\
	\end{aligned}
\]
and for Counting expressions
\[
	\begin{aligned}
		\cval_w(k, j) & \ldef k \\
		\cval_w(C, j) & \ldef \cval_w(e_C, j) \\
		\cval_w(\# e, j) & \ldef \sum_{i \leq j} \cval_w(e\ ?\ 1 : 0, i) \\
		\cval_w(\#_{r_s, r_e} e, j)
			& \ldef \sum_{j - r_e \leq i \leq j - r_s} \cval_w(e\ ?\ 1 : 0, i) \\
		\cval_w(e_1 + e_2, j) & \ldef \cval_w(e_1, j) + \cval_w(e_2, j) \\
		\cval_w(e_1 - e_2, j) & \ldef \cval_w(e_1, j) - \cval_w(e_2, j) \\
		\cval_w(\min(e_1, e_2), j)
			& \ldef \min(\cval_w(e_1, j), \cval_w(e_2, j)) \\
		\cval_w(\max(e_1, e_2), j)
			& \ldef \max(\cval_w(e_1, j), \cval_w(e_2, j)) \\
		\cval_w(e_1\ ?\ e_2 : e_3, j) & \ldef \begin{cases}
			 \cval_w(e_2, j) & \text{if $\cval_w(e_1, j) = \top$} \\
			 \cval_w(e_3, j) & \text{otherwise.}
		\end{cases}
	\end{aligned}
\]

\subsection{Additional Definitions for Lustre}

\subsubsection{Definition of Expression Evaluation}
\label{app:lustre}
We define $\lval(e, i)$ in full for Lustre expressions.
For Boolean expressions
\[
\begin{aligned}
	\lval(\top, i) &= \top \\
	\lval(\bot, i) &= \bot \\
	\lval(\neg e, i) &= \neg \lval(e, i) \\
	\lval(e_1 \diamond_b e_2, i)
	&= \lval(e_1, i) \diamond_b \lval(e_2, i) \\
	\lval(e_1 \bowtie_b, i)
	&= \lval(e_1, i) \bowtie_b \lval(e_2, i) \\
	\lval(\lpre(e), i) &= \begin{cases}
		\lval(e, i - 1) & \text{when $i > 0$} \\
		\text{undefined} & \text{otherwise}
	\end{cases} \\
	\lval(e_1 \lnext e_2, i) &= \begin{cases}
		\lval(e_1, i) & \text{if $i = 0$} \\
		\lval(e_2, i) & \text{otherwise}
	\end{cases}
\end{aligned}
\]
and for integer expressions
\[
\begin{aligned}
	\lval(k, i) &= k \\
	\lval(e_1 \diamond_c e_2, i)
	&= \lval(e_1, i) \diamond_c \lval(e_2, i) \\
	\lval(\lif e_1 \lthen e_2 \lelse e_3, i) &= \begin{cases}
		\lval(e_2, i) & \text{if $\lval(e_1, i) = \top$} \\
		\lval(e_3, i) & \text{otherwise} \\
	\end{cases} \\
	\lval(\lpre(e), i) &= \lval(e, i - 1) \\
	\lval(e_1 \lnext e_2, i) &= \begin{cases}
		\lval(e_1, i) & \text{if $i = 0$} \\
		\lval(e_2, i) & \text{otherwise.}
	\end{cases}
\end{aligned}
\]

\subsubsection{Translating C-RASP to Lustre}
\label{app:trans}
We define $T(e)$ in full.
For Boolean expressions (recalling that $I = a$ denotes $I = \latoi(a)$)
\[
\begin{aligned}
	T(\top) & \ldef \top \\
	T(\bot) & \ldef \bot \\
	T(a) & \ldef I = a \\
	T(B) & \ldef V_B \\
	T(\neg e) & \ldef \neg T(e) \\
	T(e_1 \diamond e_2) & \ldef T(e_1) \diamond T(e_2) \\
	T(e_1 \bowtie e_2) & \ldef T(e_1) \bowtie T(e_2) \\
	T(m\%o) & \ldef P \lmod m = o \\
\end{aligned}
\]
where $P$ is a fresh integer variable encoding the current position.
It uses an auxiliary equation $P \ldef 0 \lnext \lpre(P) + 1$.
For counting expressions
\[
\begin{aligned}
	T(k) & \ldef k \\
	T(C) & \ldef V_C \\
	T(\#(e)) & \ldef C_{\#(e)} \\
	T(\#_{r_s, r_e}(e)) & \ldef C_{\#_{r_s, r_e}(e)} \\
	T(e_1 + e_2) & \ldef T(e_1) + T(e_2) \\
	T(e_1 - e_2) & \ldef T(e_1) - T(e_2) \\
	T(\min(e_1, e_2)) & \ldef
	\lif C_{e_1} < C_{e_2} \lthen C_{e_1} \lelse C_{e_2} \\
	T(\max(e_1, e_2)) & \ldef
	\lif C_{e_2} < C_{e_1} \lthen C_{e_1} \lelse C_{e_2} \\
	T(e_1\ ?\ e_2\ :\ e_3) & \ldef
	\lif T(e_1) \lthen T(e_2) \lelse T(e_3) \\
\end{aligned}
\]
where $C_{\#(e)}$, $C_{\#_{r_s, r_e}(e)}$, $C_{e_1}$, and $C_{e_2}$ are fresh
integer variables defined via the auxiliary equations.

The additional auxiliary equations are: for each $C_{e_i}$ add the rule
\[
C_{e_i} \ldef T(e_i).
\]

\subsubsection{Additional Verification Properties}

\paragraph{Universality}

Let $P = (R_1, \ldots, R_k)$ be a C-RASP program.
Universality can be checked by inspecting the value of $R_k = B \ldef e$
(corresponding to the final C-RASP rule).
For universality, we require that $R_k$ always evaluates to $\top$.
It requires the following $\lcheck$ constraint which asserts that after the
first position, either the input $I$ is incorrect, the end of the stream marker
has not been reached, or the value of $R_k$ is $\top$.
\[
\lcheck \top \lnext (\neg(B_{\hat{I}} \land I = \leos) \lor \lpre(V_B))
\]

We define $\tvalid(P)$ to be the Lustre program with the equations $\mathcal{R}_I
\cup \bigcup_i T(R_i)$ and the check statement defined above.

\begin{proposition}
	A C-RASP program $P$ accepts all words $w \in \Sigma^+$ if $\tvalid(P)$ is
	valid.
\end{proposition}

\paragraph{Language Equality}
We can check the equivalence of two C-RASP programs $P_1$ and $P_2$ using a similar encoding to language inclusion.
We need to adjust the $\lcheck$ equation as shown below.
\[
\lcheck \top \lnext \left(
\begin{array}{c}
	\neg(B_{\hat{I}} \land I = \leos) \\
	\lor\ (\lpre(V_1) \land \lpre(V_2)) \\
	\lor\ (\neg \lpre(V_1) \land \neg \lpre(V_2))
\end{array}
\right)
\]

We define $\tequal(P_1, P_w)$ to be the Lustre program with the equations
$\mathcal{R}_I \cup \bigcup_i T(R^1_i) \cup \bigcup_i T(R^2_i)$ and the check
statement defined above.

\begin{proposition}
	A C-RASP program $P_1$ accepts the same words $w \in \Sigma^+$ accepted by a
	C-RASP program $P_2$ if $\tequal(P_1, P_2)$ is valid.
\end{proposition}
\section{Additional Details for Experiments}
\label{app:benchmark-table}

\subsection{Additional Supporting Results}
Table~\ref{tab:all-results} shows the C-RASP synthesis results for languages beyond C-RASP's
expressive power, including Parity, Tomita~3, Tomita~5, and Tomita~6. As expected, none of
these languages could be synthesized with 100\% accuracy within the 300-second timeout,
with the best training accuracy remaining below 84\%. Consequently, program minimization
and constraint learning could not be applied, as they require a fully accurate program
as a starting point.

\begin{table*}[htbp]
	\centering
	\caption{C-RASP synthesis results for languages beyond C-RASP's expressive power}
	
	\label{tab:fail-results}
	{
		\renewcommand{\arraystretch}{0.75}
		\setlength{\tabcolsep}{3pt}
		\resizebox{\linewidth}{!}{%
			\begin{tabular}{lcccccc ccc}
				\toprule
				& \multicolumn{3}{c}{\textbf{SL Synthesis}}
				& \multicolumn{3}{c}{\textbf{Program Minimization}}
				& \multicolumn{3}{c}{\textbf{Constraint Learning}} \\
				\cmidrule(lr){2-4} \cmidrule(lr){5-7} \cmidrule(lr){7-10}
				\textbf{Language}
				& \textbf{TA (\%)}			& \textbf{EA (\%)} & \textbf{Time (s)}
				& \textbf{R} & \textbf{Synth (s)} & \textbf{Verif (s)}
				& \textbf{R} & \textbf{Synth (s)} & \textbf{Verif (s)} \\
				\midrule

				Parity               & 83.12  & 82.50 & 185.63 & - & - & - & - & - & -  \\[0.10ex]
				Tomita 3            & 79.25 & 78.00 & 228.94& - & - & - & - & - & -  \\[0.10ex]
				Tomita 5        & 79.44 & 71.70 & 112.23   & - & - & - & - & - & -  \\[0.10ex]
				Tomita 6             & 75.75 & 71.48 &  183.61   & - & - & - & - & - & -  \\[0.10ex] 
				\bottomrule
			\end{tabular}%
		}
	}
\end{table*}

Table~\ref{tab:comparison} reports transformer training results across all benchmark tasks,
evaluated at three length bins: lengths $\leq 50$ (Val$_0$, training distribution),
$[51, 100]$ (Val$_1$), and $[101, 150]$ (Val$_2$, generalization). Despite the
substantial training time --- often exceeding 20 minutes and up to nearly 49 minutes
for some tasks --- transformers still struggle to generalize to longer sequences, with
Val$_2$ accuracy frequently dropping significantly compared to Val$_0$. Tasks beyond
C-RASP's expressive power (e.g., Parity, Tomita-3, Tomita-5, Tomita-6) exhibit
consistently low accuracy across all bins, while tasks within C-RASP's expressivity
(e.g., \texttt{sort}, \texttt{bin\_majority}, \texttt{majority}) achieve near-perfect
accuracy even on longer sequences.
 
\begin{table}[ht]
	\centering
	\caption{Transformer training validation accuracy (\%) at lengths $\leq 50$ (Val$_0$, training),
		$[51, 100]$ (Val$_1$), and $[101, 150]$ (Val$_2$, generalization), and training time (min) for all tasks.}
	\label{tab:comparison}
	\begin{tabular}{lcccc}
		\toprule
		\textbf{Task} & \textbf{$Val_0$} & \textbf{$Val_1$} & \textbf{$Val_2$} & \textbf{Time (min)} \\
		\midrule
		012Star\_0\_2Star              & 82.2  & 58.2  & 42.8  & 20.88 \\[1.2ex]
		AAAAStar                       & 74.7  & 73.3  & 63.4  & 48.93 \\[1.2ex]
		AAStar                         & 69.4  & 54.6  & 53.2  & 17.74 \\[1.2ex]
		AAStarBBStarCCStarDDStarEEStar & 97.4  & 71.5  & 53.6  & 24.82 \\[1.2ex]
		ABABStar                       & 75.1  & 75.0  & 75.0  & 48.34 \\[1.2ex]
		ABStar\_D\_BCStar              & 94.8  & 62.0  & 42.5  & 16.19 \\[1.2ex]
		D-12                           & 81.9  & 82.2  & 74.2  & 10.05 \\[1.2ex]
		D-2                            & 82.5  & 75.5  & 66.8  &  4.85 \\[1.2ex]
		D-3                            & 87.7  & 89.2  & 79.6  & 10.26 \\[1.2ex]
		D-4                            & 92.6  & 76.2  & 68.1  & 10.33 \\[1.2ex]
		Tomita-1                       & 97.9  & 86.8  & 84.7  &  6.89 \\[1.2ex]
		Tomita-2                       & 98.1  & 91.6  & 90.9  &  6.73 \\[1.2ex]
		Tomita-3                       & 56.8  & 57.1  & 56.7  & 21.58 \\[1.2ex]
		Tomita-4                       & 83.9  & 82.7  & 82.9  & 16.85 \\[1.2ex]
		Tomita-5                       & 72.9  & 74.2  & 74.1  & 21.19 \\[1.2ex]
		Tomita-6                       & 65.5  & 64.7  & 64.7  & 38.52 \\[1.2ex]
		Tomita-7                       & 66.4  & 49.8  & 49.9  & 17.15 \\[1.2ex]
		bin\_majority                  & 100.0 & 100.0 & 99.9  &  0.69 \\[1.2ex]
		bin\_majority\_interleave      & 100.0 &  95.9 & 70.3  &  2.82 \\[1.2ex]
		majority                       & 100.0 &  99.6 & 97.7  &  3.24 \\[1.2ex]
		parity                         &  85.1 &  62.3 & 54.2  & 13.75 \\[1.2ex]
		repeat\_copy                   & 100.0 &  32.6 &  0.0  & 10.29 \\[1.2ex]
		sort                           & 100.0 & 100.0 & 100.0 &  5.72 \\[1.2ex]
		unique\_copy                   & 100.0 &  99.9 & 98.9  &  8.95 \\[1.2ex]
		\bottomrule
	\end{tabular}
\end{table}

\subsection{CRASP Programs}

We present the explicit C-RASP programs for all benchmark languages used in our
experiments. The programs cover a diverse range of language classes, including
regular languages (Tomita~1, 2, 4, 7), context-free languages (Dyck-1,
$a^nb^nc^n$), piecewise testable languages (PT-2, PT-3, PT-5), and other
languages expressible in C-RASP (Majority, $(aa)^*$, $a^*b^*$,
$\Sigma^*ab\Sigma^*$, Existential, Next(Argmax)). Each program is presented
alongside a formal description of the language it recognizes, illustrating
the expressiveness and compactness of C-RASP as a formalism for
transformer-recognizable languages.

\paragraph{Dyck-1} is the language of well-balanced parentheses over
$\Sigma=\{l,r\}$: every prefix contains at least as many $l$ as $r$, and the
total numbers of $l$ and $r$ are equal.

\begin{tcolorbox}[
	enhanced,
	width=\linewidth,
	colback=gray!8,
	colframe=black,
	boxrule=0.6pt,
	arc=2pt,
	left=6pt,
	right=6pt,
	top=4pt,
	bottom=4pt,
	title={\(C\)-RASP program for Dyck-\(1\) over \(\Sigma=\{l,r\}\)},
	center
	]
	\[
	\begin{aligned}
		L  &= l\\
		R &= r\\
		V &= CL < CR\\
		M &= CV = C0\\
		B &= CL = CR\\
		Out &= M \land B\\
		CL &= \#l\\
		CR &= \#r\\
		CV &= \#V\\
		C0 &= 0
	\end{aligned}
	\]
\end{tcolorbox}
\paragraph{a$^*$b$^*$} is the language of strings over $\Sigma=\{a,b\}$ consisting of a block
of $a$'s followed by a block of $b$'s (possibly empty blocks).

\begin{tcolorbox}[
	enhanced,
	width=\linewidth,
	colback=gray!8,
	colframe=black,
	boxrule=0.6pt,
	arc=2pt,
	left=6pt,
	right=6pt,
	top=4pt,
	bottom=4pt,
	title={\(C\)-RASP program for $a^*b^*$  over \(\Sigma=\{l,r\}\)},
	center
	]
	\[
	\begin{aligned}
		Qa &= a\\
		Qb &= b\\
		V  &= Qa \land (0 < Cb) \\    
		Y  &= CV = 0             \\
		Cb &= \#Qb\\               
		CV &= \#V    \\   
		Out &= Y  
	\end{aligned}
	\]
\end{tcolorbox}

\paragraph{MAJORITY} is the language of strings over $\Sigma=\{a,b\}$ with at least
as many $a$ symbols as $b$ symbols.

\begin{tcolorbox}[
	enhanced,
	width=\linewidth,
	colback=gray!8,
	colframe=black,
	boxrule=0.6pt,
	arc=2pt,
	left=6pt,
	right=6pt,
	top=4pt,
	bottom=4pt,
	title={\(C\)-RASP program for Majority over \(\Sigma=\{a,b\}\)},
	center
	]
	\[
	\begin{aligned}		
		Pa &= a\\
		Pb &= b\\
		Out &= Ca \leq Cb\\
		Ca &= \#Pa\\
		Cb &= \#Pb\\
	\end{aligned}
	\]
\end{tcolorbox}

\paragraph{(aa)$^*$} is the language of even-length strings over $\Sigma=\{a\}$.

\begin{tcolorbox}[
	enhanced,
	width=\linewidth,
	colback=gray!8,
	colframe=black,
	boxrule=0.6pt,
	arc=2pt,
	left=6pt,
	right=6pt,
	top=4pt,
	bottom=4pt,
	title={\(C\)-RASP program for $(aa)^*$ over \(\Sigma=\{a\}\)},
	center
	]
	\[
	\begin{aligned}		
		CnotA &= \#(\neg a) \\
		Out &= (\%2=1) \land (CnotA=0)
	\end{aligned}
	\]
\end{tcolorbox}

\paragraph{a$^n$b$^n$c$^n$} is the language $\{a^n b^n c^n \mid n\ge 0\}$ over $\Sigma=\{a,b,c\}$.

\begin{tcolorbox}[
	enhanced,
	width=\linewidth,
	colback=gray!8,
	colframe=black,
	boxrule=0.6pt,
	arc=2pt,
	left=6pt,
	right=6pt,
	top=4pt,
	bottom=4pt,
	title={\(C\)-RASP program for $a^nb^nc^n$ over \(\Sigma=\{a, b, c\}\)},
	center
	]
	\[
	\begin{aligned}		
		Qa &= a\\
		Qb &= b\\
		Qc &= c\\
		Ca &= \#Qa\\
		Cb &= \#Qb\\
		Cc &= \#Qc\\
		A  &= Cb + Cc = 0\\
		B  &= Cc = 0\\
		QaA &= Qa \land A\\
		QbB &= Qb \land B\\
		CA &= \#QaA\\
		CB &= \#QbB\\
		Ga &= CA = Ca\\
		Gb &= CB = Cb\\
		Gabc &= (Ca = Cb) \land (Cb = Cc)\\
		Out &= Ga \land Gb \land Gabc
	\end{aligned}
	\]
\end{tcolorbox}

\paragraph{$\Sigma^*$ab$\Sigma^*$}over $\Sigma=\{a,b\}$ is the language of strings containing
the substring $ab$ at some position.

\begin{tcolorbox}[
	enhanced,
	width=\linewidth,
	colback=gray!8,
	colframe=black,
	boxrule=0.6pt,
	arc=2pt,
	left=6pt,
	right=6pt,
	top=4pt,
	bottom=4pt,
	title={\(C\)-RASP program for $\Sigma^*$ab$\Sigma^*$  over \(\Sigma=\{a, b\}\)},
	center
	]
	\[
	\begin{aligned}		
		CaPre &= \#<(1) a\\
		PaPre &= 1 \leq CaPre\\
		Qab &= b \land PaPre\\
		Cab &= \#Qab\\
		Out &= 1 \leq Cab
	\end{aligned}
	\]
\end{tcolorbox}

\paragraph{Existential} is the language of strings over $\Sigma=\{a,b\}$ containing
at least one occurrence of $a$.

\begin{tcolorbox}[
	enhanced,
	width=\linewidth,
	colback=gray!8,
	colframe=black,
	boxrule=0.6pt,
	arc=2pt,
	left=6pt,
	right=6pt,
	top=4pt,
	bottom=4pt,
	title={\(C\)-RASP program for Existential over \(\Sigma=\{a,b\}\)},
	center
	]
	\[
	\begin{aligned}		
		A &= a \\
		B &= b \\
		C1 &= \#A \\ 
		C2 &= \#B \\ 
		Out &= 1 <= C2
	\end{aligned}
	\]
\end{tcolorbox}

\paragraph{Next(Argmax)} is the language over $\Sigma$ in which, for each position
$i>1$, the observed successor $x_i$ maximises (ties allowed) the empirical
count of transitions from the previous symbol $x_{i-1}$, compared against all
alternative successors $c\in\Sigma$, using statistics computed from the prefix
up to position $i-1$.

\begin{tcolorbox}[
	enhanced,
	width=\linewidth,
	colback=gray!8,
	colframe=black,
	boxrule=0.6pt,
	arc=2pt,
	left=6pt,
	right=6pt,
	top=4pt,
	bottom=4pt,
	title={\(C\)-RASP program for Next(Argmax) over \(\Sigma=\{a, b, c\}\)},
	center
	]
	\[
	\begin{aligned}		
		C1 &= \#(1,1) a\\
		B1 &= 1 \leq C1\\
		B2 &= a \land B1\\
		B3 &= b \land B1\\
		B4 &= c \land B1\\
		C2 &= \#B2\\
		C3 &= \#B3\\
		C4 &= \#B4\\
		B5 &= C3 \leq C2\\
		B6 &= C4 \leq C2\\
		B7 &= 1 \leq (C2 + C3 + C4)\\
		B8 &= a \land B7 \land B5 \land B6\\
		Out &= 1 \leq \#B8
	\end{aligned}
	\]
\end{tcolorbox}

\paragraph{Tomita 1} is the regular language $a^*$ over $\Sigma=\{a,b\}$, i.e.,
strings containing no occurrences of $b$.

\begin{tcolorbox}[
	enhanced,
	width=\linewidth,
	colback=gray!8,
	colframe=black,
	boxrule=0.6pt,
	arc=2pt,
	left=6pt,
	right=6pt,
	top=4pt,
	bottom=4pt,
	title={\(C\)-RASP program for Tomita 1 over \(\Sigma=\{a, b\}\)},
	center
	]
	\[
	\begin{aligned}		
		B1 &= b \\
		C1 &= \#B1 \\
		Out &= (C1 = 0)
	\end{aligned}
	\]
\end{tcolorbox}

\paragraph{Tomita 2} is the regular language $(ab)^*$ over $\Sigma=\{a,b\}$, i.e.,
strings of even length that alternate starting with $a$ and ending with $b$.

\begin{tcolorbox}[
	enhanced,
	width=\linewidth,
	colback=gray!8,
	colframe=black,
	boxrule=0.6pt,
	arc=2pt,
	left=6pt,
	right=6pt,
	top=4pt,
	bottom=4pt,
	title={\(C\)-RASP program for Tomita 2 over \(\Sigma=\{a, b\}\)},
	center
	]
	\[
	\begin{aligned}		
		B1 &= (a \land \%2=0) \lor (b \land \%2=1)\\
		B2  &= \#(\neg B1)\\
		Out &= (\%2=1) \land (B2 = 0)
	\end{aligned}
	\]
\end{tcolorbox}

\paragraph{Tomita 4} is the regular language over $\Sigma=\{a,b\}$ that forbids any
substring of the form $a\,b^{2n+1}\,a^{2m+1}\,b$ for $n,m\ge 0$.
Equivalently, $w \in \textsc{Tomita-4}$ iff no occurrence of $a$ is followed by
an odd run of $b$'s, then an odd run of $a$'s, and then a $b$.

\begin{tcolorbox}[
	enhanced,
	width=\linewidth,
	colback=gray!8,
	colframe=black,
	boxrule=0.6pt,
	arc=2pt,
	left=6pt,
	right=6pt,
	top=4pt,
	bottom=4pt,
	title={\(C\)-RASP program for Tomita 4 over \(\Sigma=\{a, b\}\)},
	center
	]
	\[
	\begin{aligned}		
		A1 &= \#(1,1) a\\
		P1 &= 1 \leq A1\\
		A2 &= \#(2,2) a\\
		P2 &= 1 \leq A2\\
		AAApos &= a \land P1 \land P2\\
		CAAA &= \#AAApos\\
		Out &= (CAAA = 0)
	\end{aligned}
	\]
\end{tcolorbox}

\paragraph{Tomita 7} is the regular language $a^*b^*a^*b^*$ over $\Sigma=\{a,b\}$,
i.e., strings that can be partitioned into at most four consecutive blocks,
alternating between $a$'s and $b$'s starting with $a$.
\begin{tcolorbox}[
	enhanced,
	width=\linewidth,
	colback=gray!8,
	colframe=black,
	boxrule=0.6pt,
	arc=2pt,
	left=6pt,
	right=6pt,
	top=4pt,
	bottom=4pt,
	title={\(C\)-RASP program for Tomita 7 over \(\Sigma=\{a, b\}\)},
	center
	]
	\[
	\begin{aligned}		
		B &= b\\
		A &= a\\
		Cb    &= \#B\\
		B1 &= 1 \leq Cb\\
		B2 &= A \land B1\\
		C1  &= \#B2\\
		B3 &= 1 \leq C1\\
		B4 &= B \land B3\\
		CBAB            &= \#B4\\
		B1AB          &= 1 \leq CBAB\\
		B5 &= A \land B1AB\\
		CBadA   &= \#B5\\
		Out &= (CBadA = 0)
	\end{aligned}
	\]
\end{tcolorbox}

\paragraph{PT-2}  over $\Sigma=\{a,b\}$
consists of strings in which an $a$ occurs before some later $b$.

\begin{tcolorbox}[
	enhanced,
	width=\linewidth,
	colback=gray!8,
	colframe=black,
	boxrule=0.6pt,
	arc=2pt,
	left=6pt,
	right=6pt,
	top=4pt,
	bottom=4pt,
	title={\(C\)-RASP program for PT-2 over \(\Sigma=\{a, b\}\)},
	center
	]
	\[
	\begin{aligned}		
		A &= a\\
		B &= b\\
		CA &= \#A\\
		L1 &= 1 \leq CA\\
		QB &= B \land L1\\
		CB &= \#QB\\
		L2 &= 1 \leq CB \\
		Out &= L2
	\end{aligned}
	\]
\end{tcolorbox}

\paragraph{PT-3} over $\Sigma=\{a,b,c\}$ in which
$a$, then $b$, then $c$ occur in this order (not necessarily consecutively).
Formally, $w \in \textsc{PT-3}$ iff there exist indices $i<j<k$ such that
$w[i]=a$, $w[j]=b$, and $w[k]=c$.

\begin{tcolorbox}[
	enhanced,
	width=1\linewidth,
	colback=gray!8,
	colframe=black,
	boxrule=0.6pt,
	arc=2pt,
	left=6pt,
	right=6pt,
	top=4pt,
	bottom=4pt,
	title={\(C\)-RASP program for PT-3 over \(\Sigma=\{a, b, c\}\)},
	center
	]
	\[
	\begin{aligned}		
		A &= a\\
		B &= b\\
		C &= c\\
		CA &= \#A\\
		L1 &= 1 \leq CA\\
		QB &= B \land  L1\\
		CB &= \#QB\\
		L2 &= 1 \leq CB\\ 
		QC &= C \land  L2\\
		CC &= \#QC\\
		L3 &= 1 \leq CC\\  
		Out &= L3
	\end{aligned}
	\]
\end{tcolorbox}

\paragraph{PT-5} is the piecewise testable language over $\Sigma=\{a,b,c,d,e\}$ in
which $a,b,c,d,e$ occur in this order as a (not necessarily contiguous)
subsequence. Formally, $w \in \textsc{PT-5}$ iff there exist
$i_1<i_2<i_3<i_4<i_5$ with $w[i_1]=a$, $w[i_2]=b$, $w[i_3]=c$, $w[i_4]=d$,
and $w[i_5]=e$.

\begin{tcolorbox}[
	enhanced,
	width=\linewidth,
	colback=gray!8,
	colframe=black,
	boxrule=0.6pt,
	arc=2pt,
	left=6pt,
	right=6pt,
	top=4pt,
	bottom=4pt,
	title={\(C\)-RASP program for PT-5 over \(\Sigma=\{a, b, c, d, e\}\)},
	center
	]
	\[
	\begin{aligned}		
		A &= a\\
		B &= b\\
		C &= c\\
		D &= d\\
		E &= e\\
		CA &= \#A\\
		L1 &= 1 <= CA\\
		QB &= B \land  L1\\
		CB &= \#QB\\
		L2 &= 1 \leq CB\\
		QC &= C \land  L2\\
		CC &= \#QC\\
		L3 &= 1 \leq CC\\
		QD &= D \land L3\\
		CD &= \#QD\\
		L4 &= 1 \leq CD\\
		QE &= E \land  L4\\
		CE &= \#QE\\
		L5 &= 1 \leq CE\\
		Out &= L5
	\end{aligned}
	\]
\end{tcolorbox}

\end{document}